\documentclass[runningheads]{llncs}
\usepackage{marvosym}
\usepackage{amsmath, amssymb}
\usepackage{xcolor}
\usepackage{array}
\usepackage{colortbl}
\usepackage{multirow}
\usepackage{booktabs} 
\usepackage[T1]{fontenc}

\newcommand{\dataset}{RefPath}
\usepackage{graphicx}
\usepackage{subcaption}
\usepackage{graphicx,verbatim}
\begin{document}
\title{PathVG: A New Benchmark and Dataset for Pathology Visual Grounding}
%

\author{Chunlin Zhong\inst{1}\textsuperscript{$\star$}, Shuang Hao\inst{1}\textsuperscript{$\star$}, Junhua Wu\inst{2}\textsuperscript{$\star$}, Xiaona Chang\inst{2}\textsuperscript{$\star$}, Jiwei Jiang\inst{1}, Xiu Nie\inst{2}\textsuperscript{\Letter}, He Tang\inst{1}\textsuperscript{\Letter}, Xiang Bai\inst{1}\textsuperscript{\Letter}} 
\authorrunning{Chunlin Zhong et al.}
\institute{School of Software Engineering, Huazhong University of Science and Technology,
Wuhan 430074, China \and
Department of Pathology, Union Hospital, Tongji Medical College, Huazhong
University of Science and Technology, Wuhan 430022, China\\
    \let\thefootnote\relax\footnotetext{\textsuperscript{$\star$} Equal contribution \quad \textsuperscript{\Letter} Corresponding author}
}

\maketitle              
\begin{abstract}
With the rapid development of computational pathology, many AI-assisted diagnostic tasks have emerged. Cellular nuclei segmentation can segment various types of cells for downstream analysis, but it relies on predefined categories and lacks flexibility. Moreover, pathology visual question answering can perform image-level understanding but lacks region-level detection capability. 
To address this, we propose a new benchmark called Pathology Visual Grounding (PathVG), which aims to detect regions based on expressions with different attributes. To evaluate PathVG, we create a new dataset named RefPath which contains 27,610 images with 33,500 language-grounded boxes. Compared to visual grounding in other domains, PathVG presents pathological images at multi-scale and contains expressions with pathological knowledge. In the experimental study, we found that the biggest challenge was the implicit information underlying the pathological expressions. Based on this, we proposed Pathology Knowledge-enhanced Network (PKNet) as the baseline model for PathVG. PKNet leverages the knowledge-enhancement capabilities of Large Language Models (LLMs) to convert pathological terms with implicit information into explicit visual features, and fuses knowledge features with expression features through the designed Knowledge Fusion Module (KFM).
The proposed method achieves state-of-the-art performance on the PathVG benchmark. 

\keywords{Pathology Visual Grounding  \and Vision-Language Model \and Large Language Model.}

\end{abstract}
\section{Introduction}

\begin{figure*}[t]
  \centering
  \includegraphics[width=1\linewidth, keepaspectratio]{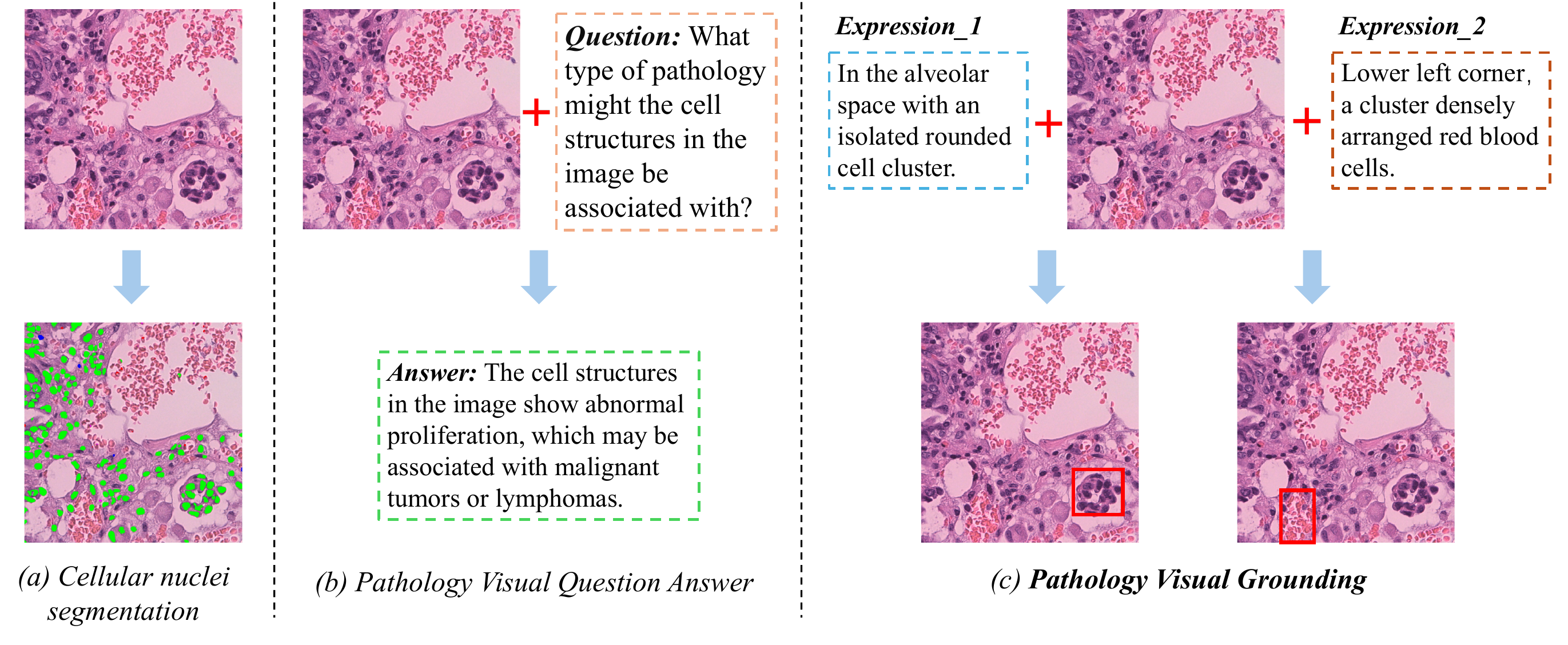}

  \caption{A comparison of (a) Cellular nuclei segmentation, (b) Pathology Visual Question Answer and (c) our proposed PathVG benchmark.}
  \label{fig1}
\end{figure*}

Pathology is the cornerstone of modern medicine, playing a crucial role in disease diagnosis and understanding. With the development of artificial intelligence, computational pathology has made significant strides, such as whole-slide cancer subtyping and survival prediction~\cite{class_1,class_2,class_3}, cellular nuclei segmentation~\cite{segmentation_1}, and pathology visual question answering~\cite{PathMMU,PathVQA}. However, current computational pathology tasks still face certain limitations. 
Cancer subtype and survival prediction are based on overall predictions from whole slide images, and patch-level analysis cannot be performed.
Cellular nuclei segmentation (Fig. \ref{fig1}(a)) can segment various types of cells for downstream analysis, but it relies on predefined categories and lacks flexibility. Moreover, pathology visual question answering (Fig. \ref{fig1}(b)) focuses on image-level understanding and cannot perform region-level detection.
In clinical practice, however, it is often necessary to detect different regions based on factors such as different organs and cancer types or in response to referring human input.
To address these challenges, we propose a novel benchmark, PathVG, which provides flexible and region-level detection capabilities.
In contrast, PathVG(Fig. \ref{fig1}(c)) allows for the localization of different regions by inputting various expressions.

\begin{figure*}[t]
  \centering
  \includegraphics[width=1\linewidth, keepaspectratio]{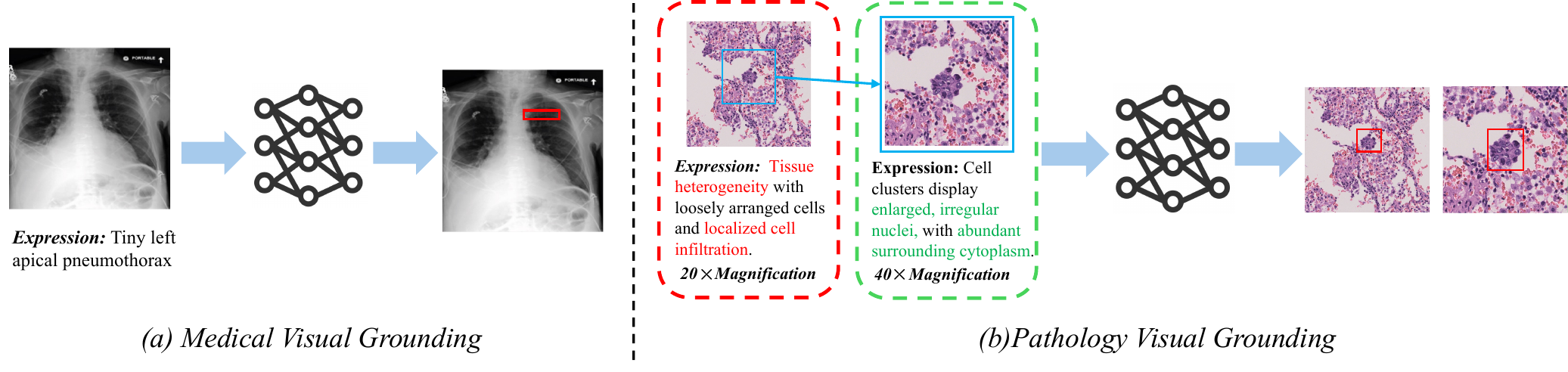}

  \caption{(a) Previous Medical Visual Grounding. (b) Pathology Visual Grounding: Identical region at Lower \textbf{(Left)} and Higher \textbf{(Right)} Magnification, with expressions for cell arrangement and interactions with neighboring cells (\textcolor{red!90}{\textbf{Red}}), as well as cell structure and growth (\textcolor{green}{\textbf{Green}}).}
  \label{fig2}
\end{figure*}

In recent years, Medical Visual Grounding (MVG) has already been explored~\cite{MX-CXR,taco,EVT,bird}. As shown in Fig.\ref{fig2} (a) and (b), due to the uniqueness of pathological data, PathVG presents two key distinctions compared to the previous MVG: (1) \textbf{Multi-scale pathological images: }A uniqueness of pathological images is that the same region exhibits different pathological features at varying magnification levels. Higher magnification images highlight cell structure and growth, while lower-magnification images reveal cell arrangement and interaction with neighboring cells. (2) \textbf{Expressions with pathological knowledge}: PathVG localizes specific regions from multiple pathological perspectives, such as cell structure, growth patterns, cell arrangement, and interactions with neighboring cells, to accurately localize regions under different magnification. To align with these two distinctions, we introduce a novel dataset called RefPath, specifically tailored for PathVG.
PathVG includes 27,610 images with 33,500 language-grounded boxes. 

Compared to visual grounding in other domains, the main challenge of PathVG lies in the implicit information underlying the pathological expressions, which makes it difficult to associate them with pathological images. The expressions in RefPath describe pathological region from multiple perspectives, requiring the model to understand a wide range of specialized terms. This becomes a challenging task for models without prior knowledge of pathology.

To address this challenge, we introduced the Pathology Knowledge-enhanced Network (PKNet), which leverages the knowledge enhancement capabilities of LLMs, transforming implicit pathological terms into explicit visual information, and better linking pathological expressions with pathological images.
Building upon this, we have designed a Knowledge Branch specifically for knowledge enhancement, as well as a Knowledge Fusion module (KFM) to better fuse knowledge and expression features. 

In summary, our contributions are listed as follows:
\begin{enumerate}
\item We propose a novel benchmark, Pathology Visual Grounding (PathVG), which enables flexible and region-level detection in pathological images.
\item We present RefPath, a large-scale dataset consisting of 27,610 images and 33,500 language-grounded boxes, tailored to the uniqueness of pathology.
\item We developed a baseline model, the Pathology Knowledge-enhanced Network (PKNet), which leverages knowledge enhancement from LLMs to transform implicit pathological expressions into explicit visual features.
\end{enumerate}
\section{RefPath: A Large-scale Dataset for PathVG}
\subsection{Dataset Collection and Annotation }
To adapt to the PathVG benchmark, we have specifically built a new large-scale dataset, RefPath.
To ensure the quality of our dataset, we have carefully devised a three-step data processing and generation protocol.
\begin{figure*}[t]
  \centering
  \includegraphics[width=1\linewidth, keepaspectratio]{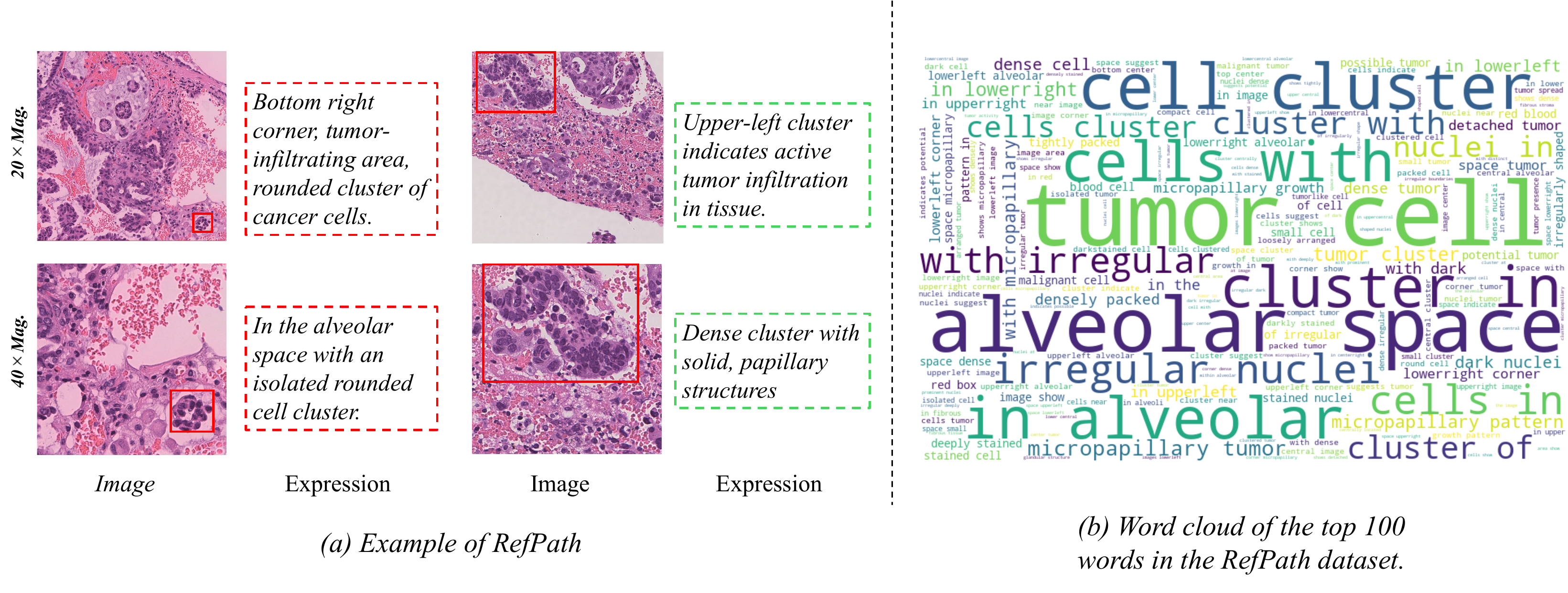}

  \caption{(a): Example of RefPath: Displayed the differences in images and at different magnification levels. (b): Word cloud of the top 100 words in the RefPath dataset. Displayed the pathological terms in the expression. }
  \label{refpath}
\end{figure*}

\textbf{Step 1: Data Collection and Preprocessing.} The first step involved collecting clinical whole-slide images, cropped at 40$\times$ and 20$\times$ magnifications with a resolution of 1024$\times$1024 pixels. This multi-scale approach captures pathological features at different magnification levels and aligns with clinical practices. Pathological experts manually labeled language-grounded boxes to identify cell clusters or regions indicative of malignancy. These annotations trained the YOLOv10~\cite{yolo} network for automated region detection. Candidate boxes were reviewed by experts to filter out irrelevant or unclear annotations, ensuring only high-quality data for downstream steps.

\textbf{Step 2: Pathological expression Generation.} In addition to language-grounded boxes annotations, pathological experts provided detailed expressions for selected images. These expression annotations served as key examples for few-shot learning by GPT-4V, improving its ability to generate accurate, context-rich pathological expressions. To optimize GPT-4V’s output, we designed specific prompts to focus the model on the morphological features of cells and tissues within the annotated regions, ensuring high precision and relevance. 

\textbf{Step 3: Expert Validation.} We divided the dataset into training and testing sets, and invited professional pathology experts to manually review the testing set. The experts evaluated the textual expression based on the following criteria: (1) whether they conform to standard clinical expression; (2) whether they correspond accurately to the specified regions; and (3) whether they provide a multifaceted, fine-grained portrayal of the specified regions. Samples failing to meet any of these criteria were deemed invalid and removed from the RefPath dataset. Ultimately, the training set comprises 24757 images with 30452 language-grounded boxes, while the testing set includes 2853 images with 3048 language-grounded boxes.
\subsection{Dataset Statistics}
As shown in Fig.\ref{refpath}, we present the uniqueness of the Refpath dataset from two aspects. First, the dataset contains multi-scale images, as shown in Fig.\ref{refpath}(a). At low(20$\times$) magnification, the images emphasize cell arrangement and interactions with neighboring cells, such as infiltration observed in the expression. At high(40$\times$) magnification, the focus shifts to cell structure and growth, such as alveolar spaces and papillary structures observed in the expression. Secondly, the Expression contains various pathology-related terms, as shown in Fig.\ref{refpath}(b). We can observe that the RefPath dataset includes pathological terms such as ‘tumor cells,’ ‘alveolar spaces,’ ‘irregular nuclei,’ and so on.
\section{Method}
\begin{figure*}[t]
  \centering
  \includegraphics[width=1\linewidth, keepaspectratio]{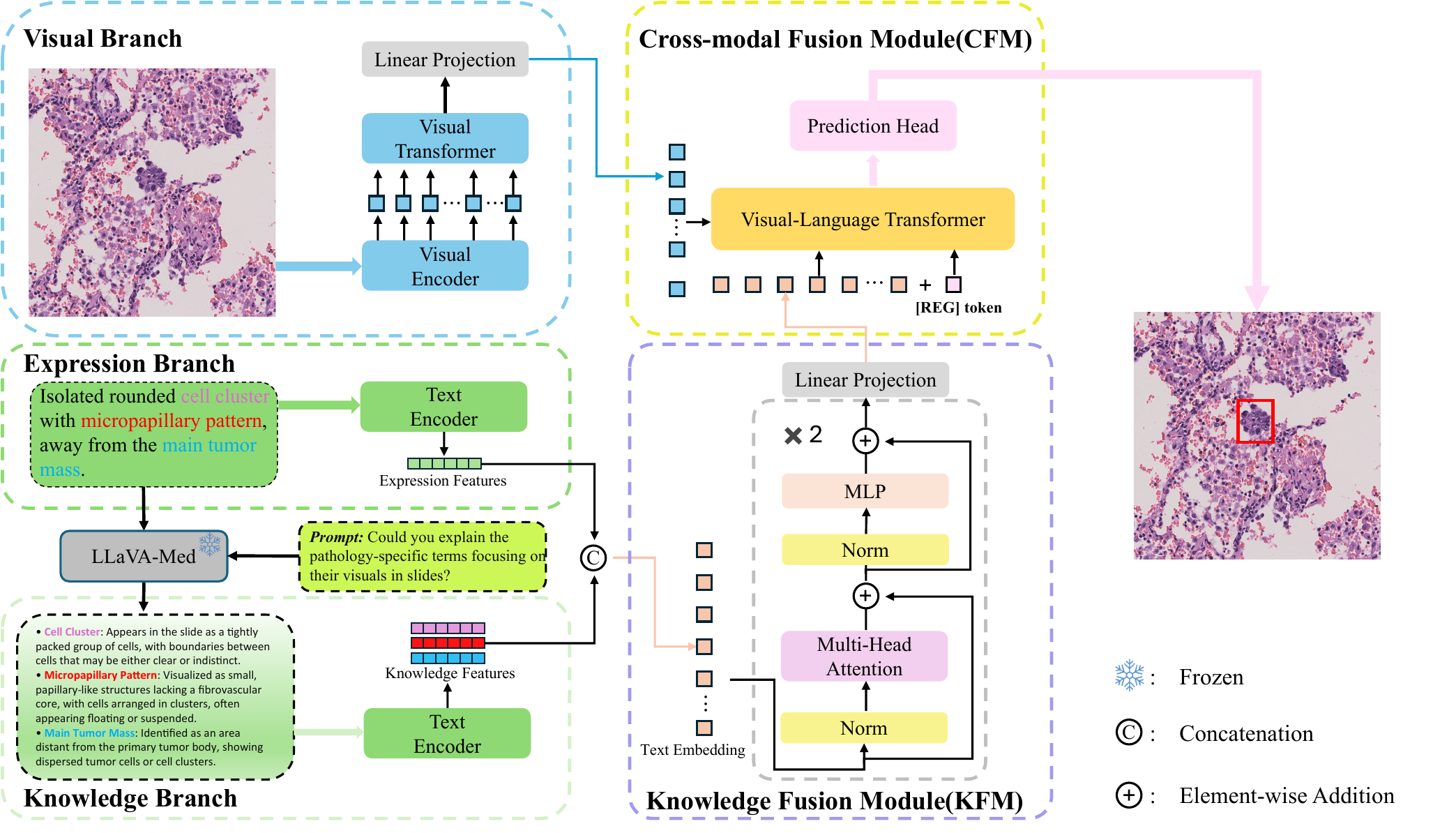}

  \caption{Overview of the proposed method. The model uses the knowledge enhancement ability of LLMs to connect pathological expressions with visual features, integrates Expression and Knowledge features through KFM, and outputs the final language-grounded boxes by combining visual features with CFM.}
  \label{fig:overall}
\end{figure*}
The PathVG problem is defined as follows: Given a image $\mathcal{I}$ along with its associated expression  \(T\), PathVG aims to output a 4-dimensional language-grounded box  \(b = (x, y, w, h)\), where\( (x, y)\) represents the center coordinates of the box and w and h denote its width and height.
\subsection{Model Architecture}
As shown in Fig.~\ref{fig:overall}, PKNet consists of five main components:(1) Visual Branch, (2) Expression Branch, (3) Knowledge Branch, (4) Knowledge Fusion Module, and (5) Cross-modal Fusion Module(CFM).\\
\textbf{Visual Branch}: Following the common practice~\cite{TransCP}, the visual encoder starts with a CNN backbone, followed by the visual transformer.We choose the ResNet-50 as the CNN backbone. The visual transformer includes 6 stacked transformer encoder layers. Each transformer encoder layer includes a multi-head self-attention layer and an FFN. 
Given an pathological image \( \mathcal{I} \in \mathbb{R}^{3 \times H \times W} \) as input of ResNet-50 to generate a 2D feature map $\mathcal{Z}\in \mathbb{R}^{C_v \times H_v \times W_v}$. The channel dimension $C_v$ is 256 and the width $W_v$ and height $H_v$ of the 2D feature map are $\frac{1}{32}$ of the original image size. Next, we flatten $\mathcal{Z}$ into $\mathcal{Z}_v \in \mathbb{R}^{C_v \times N_v} $, where $N_v = H_v*W_v$. Finally, pass $\mathcal{Z}_v$ through the transformer block to obtain the final visual feature $\mathcal{F}_v \in \mathbb{R}^{C_v \times N_v}$.
\\
\textbf{Expression Branch}:
We use the 12-layer BERT as our expression text encoder. Given an expression as the input of this branch, we first convert each word ID into a one-hot vector. Then, in the token embedding layer, we tokenize each one-hot vector into a language token. After that, we take the language tokens as inputs of the expression transformer, and generate the advanced language features \(\mathcal{F}_e \in \mathbb{R}^{C_e\times N_e}\), where \(C_e\) is the output dimension of transformer, \( N_e\) is the number of language tokens. The process is formulated as follows:
\begin{equation}
\mathcal{F}_e = \mathcal{E}_t(T,\theta_t) ,
\end{equation}
where \(\mathcal{E}_t(\cdot,\theta_t)\) is the text encoder stated above.  \(\theta_t\) denotes the encoder parameters.\\
\textbf{Knowledge Branch}: The Knowledge Branch is similar to the Expression Branch. Considering the similarity between knowledge and expression, we use the same encoder to extract knowledge features $\mathcal{F}_k$. The process is formulated as follows:
\begin{equation}
    \mathcal{F}_k = \mathcal{E}_t(\mathcal{H}(T,P),\theta_t) ,
\end{equation}
where \(\mathcal{H}\) represents the LLM we use to associate pathological terms with corresponding 
 visual features, and  \(P \) is the prompt used in the LLM.\\
\textbf{Knowledge Fusion Module}: After the individual expression and knowledge encoding, we obtain \(\mathcal{F}_e\) and \(\mathcal{F}_k\). To integrate these two features, we propose KFM, which consists of two transformer layers. Each layer includes a multi-head self-attention layer(MSA) and a FFN. Use KFM to get the language features $\mathcal{F}_l$. The process of KFM is formulated as follows:
\begin{equation}
    \mathcal{F}_l= FFN(MSA(Concat(\mathcal{F}_e,\mathcal{F}_k)) .
\end{equation}
\textbf{Cross-modal Fusion Module}: The CFM module includes two linear projection layers (one for each modality) and a visual-language(V-L) transformer (with a stack of 6 transformer encoder layers). CFM is used to integrate the fused \(\mathcal{F}_v\in \mathbb{R}^{C_v\times N_v}\) with \(\mathcal{F}_l \in \mathbb{R}^{C_l\times N_l}\). First, the features of both modalities project to the same channel dimension through a linear projection layer. We denote the projected visual features and textual features as \(\mathcal{P}_v\in \mathbb{R}^{C_p\times N_v}\) and  \(\mathcal{P}_l\in \mathbb{R}^{C_p\times N_l}\), respectively. Then, we prepend a learnable embedding ([REG] token) join to  \(\mathcal{P}_v\in \mathbb{R}^{C_p\times N_v}\) and \(\mathcal{P}_l\in \mathbb{R}^{C_p\times N_l}\) as the input $\mathcal{X}_0$. 
After that, we input $\mathcal{X}_0$ into the V-L transformer to obtain $REG_{output}$. Finally, we leverage $REG_{output}$ from the V-L transformer as the input of our prediction head. The prediction head consists of an MLP layer. The output of it is a sequence \(b \) organized as (x, y, w, h) that means the coordinates of the top left vertex, the width, and the length for the regressed bounding box.

\subsection{Loss Function}
The model’s training uses the L1 and IoU loss functions.
 \begin{equation}
     \mathcal{L} = \lambda_{l1} \mathcal{L}_{l1}(\mathcal{P},\mathcal{GT}) +\lambda_{iou} \mathcal{L}_{iou}(\mathcal{P},\mathcal{GT}) ,
 \end{equation}
where \(\mathcal{P}\) and \(\mathcal{GT}\) denote the regressed bounding box and the ground truth bounding box, respectively. \(\lambda_{l1}\) and \(\lambda_{iou}\) are two trade-off factors that balance the two losses which are set to 5 and 2 empirically.
\section{Experiment}
\begin{table*}[t]
\centering
\caption{PathVG results on RefPath with respect to Acc and mIoU. $\uparrow$ denotes that a larger value is better. We highlight the best in the \textcolor{red!70}{\textbf{red}}. $^\dag$ represents Multimodal Large Language Model. }
\fontsize{8}{12}\selectfont
\begin{tabular}{c|c|c|cccccc}
\toprule
\multirow{2}{*}{\textbf{Model}} &\multirow{2}{*}{\textbf{Venue}}  & \multirow{2}{*}{\shortstack{\textbf{Visual/Text}\\\textbf{Encoder}}} &\multicolumn{2}{c}{\textbf{\(\dataset_{all}\)}}&\multicolumn{2}{c}{\textbf{\(\dataset_{40 \times }\)}}&\multicolumn{2}{c}{\textbf{\(\dataset_{20\times}\)}}\cr

 &&& $Acc\uparrow$ & $mIOU\uparrow$ & $Acc\uparrow$ & $mIOU\uparrow$ & $Acc\uparrow$ & $mIOU\uparrow$ \cr
 
 \hline
 TransVG\cite{TransVG} &ICCV'21& RN50/BERT-B & 58.40 &52.86&68.72 &66.75&50.29&41.94\cr
 SeqTR\cite{SeqTR}&ECCV'22 & DN53/BiGRU & 55.84 &51.96&72.65 &71.13&42.57&36.78\cr
 CLIPVG\cite{CLIPVG} &TMM'23& CLIP-B/CLIP-B &58.89 &53.97  &75.52&72.14&45.81&39.67\cr
 LLaVa-Med$^\dag$\cite{llava_med}&NeuIPS'23&CLIP-L/LLaMa&62.32  &57.96  &73.52&70.24&53.51&48.31\cr
  TransCP\cite{TransCP}&TPAMI'24 & RN50/BERT-B&61.73  &56.81  &74.27&71.92&51.87&44.93\cr
 SimVG\cite{SimVG} &NeurIPS'24& ViT-B/BERT-B &63.94&59.42 &75.36&73.18  &52.52&46.92 \cr
 D-MDETR~\cite{DMDETR}&TPAMI'24& CLIP-B/CLIP-B &64.92&57.69& 76.29 &73.10  &55.98&45.57 \cr
 \hline
 Ours&-& RN50/BERT-B&\textcolor{red!70}{\textbf{69.95}} & \textcolor{red!70}{\textbf{63.49}} & \textcolor{red!70}{\textbf{80.48}} & \textcolor{red!70}{\textbf{76.88}}&\textcolor{red!70}{\textbf{61.66}}&\textcolor{red!70}{\textbf{52.95}}\cr
 \bottomrule
\end{tabular}
\label{tab:sota}
\end{table*}
\textbf{Dataset.}
The RefPath dataset includes 27,610 images with 33,500 language-grounded boxes. The training set has 24,757 images with 30452 language-grounded boxes, while the test set contains 2,853 images with 3,048 language-grounded boxes. It is further divided into two subsets: the 40$\times$ subset with 1,342 language-grounded boxes and the 20$\times$ subset with 1,706 language-grounded boxes. All methods are evaluated on the same training and test sets.
\\
\textbf{Evaluation Metric.}
To evaluate the model’s performance on the PathVG, we follow the standard protocol for visual grounding~\cite{TransVG} to report accuracy (Acc\%). Due to the unique nature of pathological images at different magnifications, using the same IoU threshold across magnifications is unreasonable. Therefore, we set the IoU threshold to 0.7 for 40$\times$ images and 0.5 for 20$\times$ images. Additionally, we use mIoU\% for a more comprehensive comparison.
\\
\textbf{Implementation details.}
We use a single NVIDIA GeForce RTX 3090 GPU for training and testing. The weights of the CNN backbone and Transformer encoder are initialized using the pre-trained DETR model. The AdamW optimizer is employed with an initial learning rate of 1e-5. The model is trained for a total of 90 epochs. For the other comparative methods, we follow the training and testing configurations specified in their respective papers. \textit{Our base model does not use a pre-aligned vision-language model, and all the pre-trained vision encoder and text encoder have not seen pathological images.}
\begin{table*}[t]
\centering
\caption{Ablation Study. (b) refers to joining the Knowledge text and expression text into a long text, which is then input into the Expression Branch; (c) uses a Knowledge Branch to extract knowledge features; (d) builds upon (c) by adding the KFM module to fuse knowledge features and expression features. }
\fontsize{8}{12}\selectfont
\begin{tabular}{cccc|cccccc}
\toprule
&\multirow{2}{*}{\textbf{Know. Input}} &\multirow{2}{*}{\textbf{Know. Branch}} &\multirow{2}{*}{\textbf{KFM}}&\multicolumn{2}{c}{\textbf{\(\dataset_{all}\)}}&\multicolumn{2}{c}{\textbf{\(\dataset_{40\times}\)}}&\multicolumn{2}{c}{\textbf{\(\dataset_{20\times}\)}}\cr

 &&&& $Acc\uparrow$ & $mIOU\uparrow$ & $Acc\uparrow$ & $mIOU\uparrow$ & $Acc\uparrow$ & $mIOU\uparrow$ \cr
 
 \midrule
(a)&&& & 61.73  &56.81  &74.27&71.92&51.87&44.93\cr
(b)&\checkmark&  && 62.29 &58.38&74.65 &73.13&52.57&46.78\cr
(c)&\checkmark& \checkmark && 67.60 &61.03&77.69 &74.34&59.67&50.56\cr
(d)&\checkmark&\checkmark&\checkmark&\textbf{69.95} &\textbf{63.49}& \textbf{80.48}& \textbf{76.88}& \textbf{61.66 }&\textbf{52.95}\cr

 \bottomrule
\end{tabular}
\label{tab:ablation}
\end{table*}\\
\textbf{Results on PathVG.} Tab. \ref{tab:sota} presents a comparison of the performance of different models on the RefPath dataset.
The compared methods include TransVG\cite{TransVG}, SeqTR\cite{SeqTR}, CLIPVG\cite{CLIPVG},  LLava-Med\cite{llava_med}, TransCP\cite{TransCP}, SimVG\cite{SimVG} and Dynamic-MDETR\cite{DMDETR}.
For LLaVa-Med, We changed the dataset format to the Med-GRIT-270k\cite{bird} dataset to train LLaVa-med.
As can be seen, our method achieves the best performance across all evaluation metrics, especially on the more challenging $\dataset_{20\times}$. This outstanding performance can be attributed to the need for more accurate understanding of the pathological expression in lower-magnification images.
In contrast, SeqTR performs poorly primarily because its text encoder is a simple BiGRU, which limits its ability to comprehend pathological expression.\\
\textbf{Ablation Study.} We conducted ablation experiments on the Refpath to evaluate the effectiveness of each component in PKNet. Tab. \ref{tab:ablation} presents the quantitative results for each configuration. First, in the baseline setup without additional information input, as shown in (a) in Tab. \ref{tab:ablation}, the performance is suboptimal. Next, we consider incorporating knowledge information as an additional input, as shown in (b). By simply concatenating the two text as input, we observe a slight improvement, though the effect is not significant. Subsequently, to better extract both knowledge and expression information, we designed a dedicated Knowledge Branch, which led to a considerable improvement. Finally, we introduced a specially designed KFM module to fuse knowledge features and expression features, resulting in further gains and achieving the best performance.
\section{Conclusion}
In this paper, to address the limitations of existing computational pathology tasks. We propose a new benchmark, PathVG, and a dedicated dataset, RefPath. PathVG is a new benchmark that enables precise localization of specific regions in pathological images using fine-grained text descriptions. The proposed RefPath dataset contains over 27,000 images with detailed annotations. Building upon this, we introduce a new base model, PKNet, which leverages the knowledge-enhancement capabilities of large models to effectively bridge the gap between pathological expression and images.

\textbf{Limitation.} 
The benchmark and method proposed in this paper are based on fully supervised learning. However, in the medical field, obtaining annotations is costly. Therefore, we plan to explore semi-supervised and unsupervised learning approaches in future work to reduce reliance on expensive annotations.
{
    \small
    \bibliographystyle{plain}
    \bibliography{MICCAI2025_paper_template}
}
\end{document}